\documentclass{article}

\usepackage{spconf,amsmath,graphicx}
\usepackage{fancyhdr}
\usepackage{enumitem}
\setlist{nosep, leftmargin=14pt}

\usepackage{mwe} 
\usepackage{manyfoot}%
\usepackage{booktabs}%
\usepackage{algorithm}%
\usepackage{algorithmicx}%
\usepackage{algpseudocode}%
\usepackage{listings}%

\usepackage{tabularx}
\usepackage{lipsum}
\usepackage{bbding}
\usepackage{tikz}
\usepackage{amssymb}
\usepackage{multirow}
\usepackage{booktabs}
\usepackage{amssymb}
\usepackage{amsfonts}
\usepackage{url}
\usepackage[colorlinks=true,citecolor=cyan,urlcolor=blue]{hyperref}

\def\x{{\mathbf x}}
\def\L{{\cal L}}

\title{Overcoming Dimensional Collapse in Self-supervised Contrastive Learning for Medical Image Segmentation}

\name{Jamshid Hassanpour$^{\star}$ \qquad Vinkle Srivastav$^{\star \dagger}$
\qquad Didier Mutter$^{\dagger}$
\qquad Nicolas Padoy$^{\star \dagger}$}

\address{$^{\star}$ ICube, University of Strasbourg, CNRS, Strasbourg, France \\
    $^{\dagger}$IHU Strasbourg, Strasbourg, France}

\begin{document}
%

\maketitle
%



\def\x{{\mathbf x}}
\def\L{{\cal L}}

%
%
%
%
%
%
\begin{abstract}
Self-supervised learning (SSL) approaches have achieved great success when the amount of labeled data is limited. Within SSL, models learn robust feature representations by solving pretext tasks. One such pretext task is contrastive learning, which involves forming pairs of similar and dissimilar input samples, guiding the model to distinguish between them. In this work, we investigate the application of contrastive learning to the domain of medical image analysis. Our findings reveal that MoCo v2, a state-of-the-art contrastive learning method, encounters \textit{dimensional collapse} when applied to medical images. This is attributed to the high degree of inter-image similarity shared between the medical images. To address this, we propose two key contributions: \textit{local feature learning} and \textit{feature decorrelation}. Local feature learning improves the ability of the model to focus on the local regions of the image, while feature decorrelation removes the linear dependence among the features. Our experimental findings demonstrate that our contributions significantly enhance the model's performance in the downstream task of medical segmentation, both in the linear evaluation and full fine-tuning settings. This work illustrates the importance of effectively adapting SSL techniques to the characteristics of medical imaging tasks. The source code will be made publicly available at: \url{https://github.com/CAMMA-public/med-moco}
\\
\end{abstract}
\section{Introduction}

Medical image segmentation plays an important role in medical diagnosis and training workflows. Recent advances in computing power alongside the success of deep learning models in segmentation tasks have encouraged researchers to leverage artificial intelligence in this field to develop fully-supervised models for medical segmentation~\cite{isensee2019nnunet, chen2021transunet, hatamizadeh2022unetr}. Although fully supervised models in deep learning have shown promising results in segmentation, the limited amount of available labeled data poses challenges for their training. To address this issue, researchers have shown interest in developing self-supervised learning (SSL) approaches, especially when annotated data is scarce~\cite{xie2022simmim, chen2020improved}. The SSL approaches employ designated pretext tasks on the unlabeled data to learn suitable features in an unsupervised manner, which are subsequently applied to downstream tasks such as classification and segmentation. One popular pretext task is contrastive learning, which relies on learning invariance by treating individual input and its transformations as positive pairs while considering every other pair of individual inputs as negative~\cite{chen2020simple}. 

\begin{figure}[t!]
\centering
\includegraphics[width=0.45\textwidth]{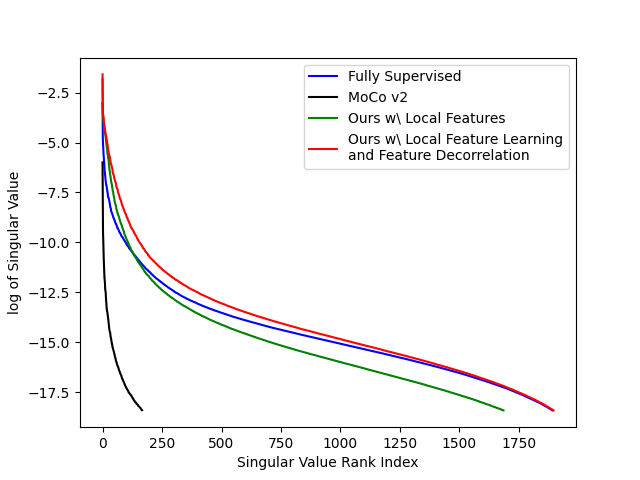}
\caption{Spectrum of singular values in different pre-training schemes. As shown, the singular values in the regular MoCo v2 pre-training are near zero and cause the dimensional collapse. Local feature learning and feature decorrelation improve the backbone features and thus increase the representation dimension.}
\label{fig:svd}
\end{figure}
Several frameworks have been introduced that employ contrastive learning to learn robust representations~\cite{chen2020simple,he2020momentum, grill2020bootstrap}. These state-of-the-art methods have demonstrated success in general computer vision; however, their performance declines when confronted with medical imaging data. In contrast to the natural images, the medical images share low semantic differences and high structural similarities due to the consistent human anatomy. In this paper, we employ MoCo v2~\cite{he2020momentum, chen2020improved}, a state-of-the-art contrastive learning method, which uses data augmentations to generate positive pairs and a momentum encoder with a dynamic dictionary as a queue to generate negative pairs for contrastive learning. However, by applying MoCo v2 on the medical images, we observe \textit{dimensional collapse} where the model doesn't utilize the full potential of the representation space and restricts the representation to certain dimensions~\cite{hua2021feature}. To illustrate the dimensional collapse in the medical images, Fig.~\ref{fig:svd} shows the spectrum for Singular Value Decomposition (SVD) on the covariance matrix of the representations. If singular values reduce to zero, it conveys the occurrence of a model collapse within the respective dimension. The spectrum for the sorted singular values in logarithmic scale for the representation features reveals that the conventional training of MoCo v2 doesn't effectively utilize the underlying dimensions of backbone features, thereby making the embedding space to restrict itself to the subspace of the full representation space. 
We hypothesize that this occurs because MoCo v2 is designed to differentiate the global features of the input image through the aggregation of local anatomical region features, rather than explicitly emphasizing the need to distinguish these local anatomical regions within the learning objective. Moreover, inter-image similarities between medical input images lead to a significant correlation among the learned features, which further hinders the model's capacity to acquire meaningful representations.

\begin{figure*}[h]
\centering
\includegraphics[width=0.9\textwidth]{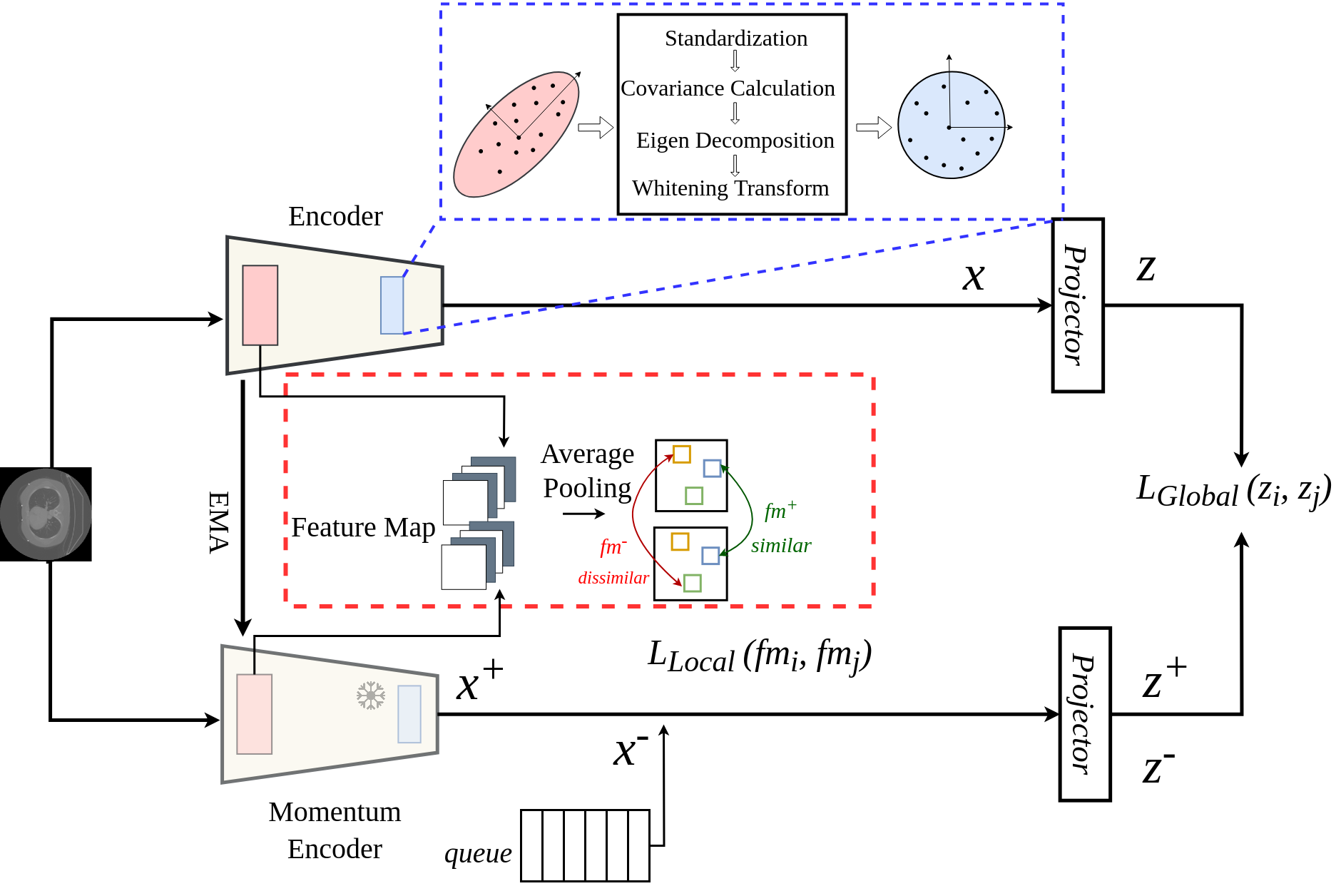}
\caption{The proposed architecture of the modified MoCo v2: The local loss is applied to the averaged feature maps from the first layer of the backbone, and ZCA whitening is done in the last layer of the backbone before the projector.}\label{fig:full_arch}
\end{figure*}

In this work, we aim to address the dimensional collapse of MoCo v2 in medical imaging and propose two contributions to enhance the feature representations. Firstly, we introduce \textit{local feature learning} that focuses on differentiating between the local regions within the input feature maps by incorporating a contrastive learning objective on the local patches of the feature maps. This helps to learn the fine-grained local features, essential for the task of medical segmentation. Secondly, we introduce \textit{feature decorrelation} that uses eigenvalue decomposition to rescale and rotate the features at the final layer of the backbone. This process enhances the model's performance by removing the correlation on the final feature map, thereby mitigating dimensional collapse during the pretraining stage. As shown in Fig.~\ref{fig:svd}, SVD analysis demonstrates that incorporating local feature learning and feature decorrelation helps in preventing the occurrence of dimensional collapse. Fig.~\ref{fig:full_arch} shows the complete pre-training pipeline.

We conduct experiments by integrating local feature learning and feature decorrelation in MoCo v2 pre-training on AbdomenCT-1K~\cite{ma2021abdomenct} dataset. Subsequently, we evaluate the pretrained backbone on Multi Atlas Labeling Beyond The Cranial Vault (BTCV) dataset~\cite{landman2015miccai} for the segmentation task. Our results demonstrate a noteworthy improvement of $8\%$ in the mean Dice Score (mDS) in the linear evaluation compared to baseline MoCo v2 training. Also, it improves the performance in the fine-tuning scheme compared to the standard training of MoCo v2. 

\section{Methodology}
To establish a baseline, we initially train MoCo v2~\cite{chen2020improved} algorithm with optimal data augmentation scheme and hyperparameters. 
In the following, we first explain the MoCo v2 method, followed by our contributions, local features learning, and feature decorrelation, to improve the learned representations. 

\subsection{MoCo v2}


MoCo v2 uses a contrastive learning objective to learn the visual representations from unlabeled input images. It does so by comparing different augmentations of an image to a repository of stored image representations, encouraging the model to pull similar input samples, called positive pairs, closer in the feature space while pushing dissimilar inputs, called negative pairs, apart. It uses two sets of neural network backbone - \textit{encoder} and \textit{momentum encoder}. The positive pairs are derived by passing different augmented input images through the \textit{encoder} and the \textit{momentum encoder} models. It then leverages the \textit{momentum encoder} to create a dynamic dictionary of features in a \textit{queue}, which serve as the negative samples. The weights of the \textit{momentum encoder} are updated using exponential moving average (\textit{EMA}) of the weights of the \textit{encoder}. The EMA ensures that the updates to the \textit{momentum encoder} are gradual and smooth, preventing drastic changes that could destabilize the learning process. It uses InfoNCE-based contrastive loss~\cite{oh2016deep} as given below:

\begin{equation}
\label{eq:infoNCE_z}
\mathcal{L}_{\text{Global}} = - \sum_{(i, j)}\log(\frac{\exp(\text{{$sim$}}(z_i, z_i^+))}{\sum_{j=1}^{N}\exp(\text{$sim$}(z_i, z_j^-))}),
\end{equation} 
where $z_i$ is the mapped representation for input image $i$ by the \textit{projector} after the encoder, $z_i, z_i^+$ are the positive pairs, $z_i, z_j^-$ are the negative pairs, and $sim$ is a similarity-based metric like cosine similarity.

\subsection{Local feature learning}

While $\mathcal{L}_{Global}$ in MoCo v2 focuses on the projected embedding at the global level, we define a loss function that directly affects the representations locally. Fig.~\ref{fig:full_arch} shows the big picture of the proposed method. 
We consider $x$ and $x^{+}$ as feature maps of the first layer of the encoder and momentum encoder, respectively. Subsequently, we sample $K$ non-overlapping patches from theses feature maps. $fm_{i}$ and $fm^{+}_{i}$ are the averaged-pooled patches located in the region $i$ to form the local positive pairs from the feature maps. The local patches at other locations, $j$, are considered negative patches, see Fig.~\ref{fig:full_arch}. We then apply the InfoNCE-based loss~\cite{oh2016deep} based on the locations of the patches. Eq.~\ref{eq:infoNCE} shows  $\mathcal{L}_{Local}$ as the similarity metric between local patches: 
\begin{equation}
\label{eq:infoNCE}
\mathcal{L}_{\text{Local}} = - \sum_{(i, j)}\log(\frac{\exp(\text{{sim}}(fm_i, fm_i^+))}{\sum_{j=1}^{K}\exp(\text{{sim}}(fm_i, fm_j^-))}).
\end{equation}

By combining global and local losses, Eq.~\ref{eq:TotalLoss} shows the overall training loss where $\lambda$ acts as a coefficient balancing the influence of the local and the global losses: 

\begin{equation}
\label{eq:TotalLoss}
\mathcal{L}_{\text{Total}} = \mathcal{L}_{Global}+\lambda \mathcal{L}_{Local}.
\end{equation}
We use the ResNet-50~\cite{he2016deep} model for the \textit{encoder} and the \textit{momentum encoder} backbones and the local loss is applied on the first layer feature maps of the backbones.
To assess the impact of the local loss, Fig.~\ref{fig:svd} shows the SVD spectrum for the validation set on the BTCV dataset~\cite{landman2015miccai}. As shown, adding local feature loss results in an increased utilization of the backbone features.

\subsection{Feature decorrelation}
For the mini-batch data $X\in{\mathbb{R}}^{d\times m}$ with size $m$,  Eq.~\ref{eq:bn} shows feature standardization done in the Batch Normalization (BN)~\cite{ioffe2015batch} layer:

\begin{equation}
\label{eq:bn}
{\mathbf{X}}_{BN}=\Sigma^{-1 / 2}\left(\mathbf{X}-\mu \cdot \mathbf{1}^T\right),
\end{equation}
 where $\mu$ and $\Sigma^{-1 / 2}$ are mean and variance for the mini-batch. Assuming covariance matrix $\Sigma = \mathbf{D}\Lambda\mathbf{D}^T$, where $\mathbf{D}$ and $\Lambda$ are eigenvectors and eigenvalues of $\Sigma$, we propose to use the ZCA whitening \cite{huang2018decorrelated} to decorrelate the input data defined as follows:
\begin{equation}
\label{eq:zca}
{\mathbf{X}}_{ZCA}=\mathbf{D} \Lambda^{-\frac{1}{2}} \mathbf{D}^T(\mathbf{X}-\mu \cdot \mathbf{1}^T).
\end{equation}

The ZCA whitening stretches and squeezes the dimension along the eigenvectors such that the eigenvalues of the covariance matrix become near $1$. This transformation decorrelates the input and normalizes its scale. We propose to use ZCA-whitening before the projector to enhance the quality of the representations. Calculating the whitening matrix $\mathbf{D} \Lambda^{-\frac{1}{2}} \mathbf{D}^T$ involves eigen decomposition and can be computationally challenging. Hence, we adopt the implementation described in~\cite{huang2019iterative}, which leverages Newton's Iteration to avoid eigen-decomposition calculation in Eq.~\ref{eq:zca}, resulting in a more efficient implementation.
We replace the last BN layer in the backbone (encoder and momentum encoder) before the projector (shown in Fig.~\ref{fig:full_arch}) with the ZCA-whitening layer to remove the correlation between the features and extend the output dimension. 
As demonstrated in Fig.~\ref{fig:svd}, decorrelating the activations further helps in preventing the model from dimensional collapse.

\section{Experiments and Results}
\subsubsection*{Datasets and data preprocessing}
For the self-supervised pre-training setup, we use AbdomenCT-1K~\cite{ma2021abdomenct} dataset and extract $215,325$ images from $1062$ patient volumes. All the input images are shuffled during the self-supervised pretraining. For downstream fully-supervised training, we evaluate the models on the Multi Atlas Labeling Beyond The Cranial Vault (BTCV) dataset~\cite{landman2015miccai}, including $30$ volumes containing $3017$ slices. The dataset encompasses $13$ abdominal organs for the segmentation task. To illustrate the effectiveness of the self-supervised pre-training in the low-labeled data regime, we use $12.5\%$, $25\%$, and $100\%$ splits of the full training set for supervised training. For $12.5\%$ and $25\%$ splits, we take three different random combinations from the full training data and show the mean and standard deviation results across the combinations. The results are reported on the validation set of the BTCV dataset due to limited access to the leaderboard evaluation.   

\subsubsection*{Implementation details}\label{sec:results}
We employ SSL and fully supervised methods from mmselfsup~\cite{mmselfsup2021} and mmseg~\cite{mmseg2020}, respectively. For our proposed model, we simplify local loss implementation by removing geometrical augmentations to ease localization in the local feature learning. We consider $K=20$ as the number of crops on ResNet-50 backbone~\cite{he2016deep}. Additionally, we replace the whitening layer in the encoder backbone with a standard BN layer for fine-tuning experiments. All backbones are initialized with weights pretrained on ImageNet. SSL models are trained with similar training hyperparameters for $200$ epochs, and fully supervised models are trained for $40K$ iterations. We evaluate the models in linear segmentation setting~\cite{bardes2022vicregl} where a linear head is attached to the backbone to match the feature map with classes. It is then upsampled through a bilinear interpolation to the original mask dimensions. We evaluate the models for the scenarios when the backbone is frozen and fine-tuned to compare the backbone capability as done in the SSL literature~\cite{he2020momentum}. Furthermore, as segmentation tasks usually use a complex decoder, we compare the models by adding a decoder from the DeepLabv3Plus model~\cite{chen2018dcan} in a full fine-tuning setting.

\subsubsection*{Results}
Table~\ref{tab:linear_frozen} shows the segmentation results for our proposed method and the baseline MoCo v2 method. Numbers are reported as the average mean Dice Score (mDS) among the splits with the standard deviation between them. Our method improves the linear segmentation results, which could be attributed to the backbone-enriched features using our proposed \emph{local feature learning} and \emph{feature decorrelation}. Furthermore, our approach shows improvement in fine-tuning evaluation with DeepLabv3Plus in comparison to MoCo v2. 
\begin{table}
    \footnotesize
    \centering
    \begin{tabular}{ccccc}
        \toprule
        \multirow{3}{*}{Split} & \multirow{3}{*}{Method} & \multicolumn{2}{c}{Linear Head} & DeepLabv3Plus \\
        \cmidrule(lr){3-4} \cmidrule(lr){5-5}
        & & Frozen & Fine-tune &  Fine-tune \\
        \midrule 
        & Fully-supervised & & $42.55\pm 2.41$  & $53.17\pm3.86$\\
        12.5\% & MoCo v2 & & $46.04\pm 4.37$  & $54.55\pm6.33$ \\
         & \textbf{Ours} &  & $\textbf{52.32}\pm{\textbf{4.75}}$ & $\textbf{56.47}\pm\textbf{4.41}$ \\
        \cmidrule(lr){2-5}
        & Fully-supervised &  & $62.95\pm 7.17$ &  $67.78\pm6.00$\\
        25\% & MoCo v2 & & $66.17\pm6.16$ & $68.56\pm6.5$  \\
        & \textbf{Ours} &  & $\textbf{67.72}\pm\textbf{5.39}$ & $\textbf{68.82}\pm\textbf{4.67}$ \\
        \cmidrule(lr){2-5}
        & Fully-supervised & $32.61$ & $63.31$ & $78.79$  \\
        100\% & MoCo v2 & $44.3$ & $78.21$ & $81.09$  \\
        & \textbf{Ours} & $\textbf{51.86}$ & $\textbf{79.38}$ & $\textbf{81.22}$ \\
        \bottomrule
    \end{tabular}
    \caption{\footnotesize{Segmentation Evaluation on BTCV dataset, numbers are Mean Dice Score and Standard Deviation in \%}}
    \label{tab:linear_frozen}
\end{table}

To gain a more in-depth understanding of the impact of \emph{local feature learning} and \emph{feature decorrelation} in the self-supervised pre-training, Table~\ref{tab:ablation} presents the results of an ablation study where we pre-train the backbone with and without \emph{local feature learning} and \emph{feature decorrelation}. The results clearly indicate that the inclusion of local features and feature decorrelation significantly enhances the model's performance in the downstream tasks. We conduct the ablation study using similar pretraining and downstream datasets. 


\begin{table}
    \footnotesize
    \centering
    \begin{tabular}{ccccccc}
        \toprule
        &Local&Feat.&\multicolumn{2}{c}{Linear Head} & DeepLabv3plus \\
        \cmidrule(lr){4-5} \cmidrule(lr){6-6}
        Method & Feat. & Decorr. & Frozen & Fine-tune &  Fine-tune \\
        \midrule 
        No SSL & & &$32.61$ & $63.31$ & $78.79$  \\
        MoCo v2 & & &$44.30$ & $78.21$ & $81.09$  \\
        \textbf{Ours} & \checkmark && $44.17$ & $78.98$ & $79.67$ \\
        \textbf{Ours} & & \checkmark & $45.27$ & $74.36$ & $81.02$ \\
        \textbf{Ours} & \checkmark& \checkmark & $\textbf{51.86}$ & $\textbf{79.38}$ & $\textbf{81.22}$ \\
        \bottomrule
    \end{tabular}
    \caption{\footnotesize{Ablation study on the pretrained backbones, numbers are Mean Dice Score in \% using $100\%$ training data of BTCV.}}
    \label{tab:ablation}
\end{table}

\section{Conclusion}
In this study, we demonstrate that state-of-the-art SSL models encounter challenges when applied to medical data, resulting in dimensional collapse. To address this issue, we introduce a novel approach by incorporating local feature learning and feature decorrelation before the projector in the MoCo v2 method. Our findings reveal that this integration effectively mitigates dimensional collapse.
Furthermore, our ablation study has shown that the incorporation of local feature learning enhances the results of fine-tuning, while feature decorrelation improves the outcomes in linear evaluation. By combining both of these methods, our proposed approach enriches the backbone features so that it achieves performance comparable to a more complex decoder in a fine-tuning setting.

\section{Acknowledgments}
\label{sec:acknowledgments}
This work was supported by French state funds within the
Investments for the future program under Grant ANR-10-
IAHU-02 (IHU Strasbourg). This work was co-funded by ArtIC project "Artificial Intelligence for Care" (grant ANR-20-THIA-0006-01) and by Région Grand Est, Inria Nancy - Grand Est, IHU of Strasbourg, University of Strasbourg and University of Haute-Alsace.

\bibliographystyle{IEEEbib}
\bibliography{strings,refs}

\begin{thebibliography}{10}

\bibitem{isensee2019nnunet}
Fabian Isensee, Paul~F Jaeger, Simon~AA Kohl, Jens Petersen, and Klaus~H
  Maier-Hein,
\newblock ``nnu-net: Self-adapting framework for u-net-based medical image
  segmentation,''
\newblock in {\em MICCAI}, 2019.

\bibitem{chen2021transunet}
Jieneng Chen, Yongyi Lu, Qihang Yu, Xiangde Luo, Ehsan Adeli, Yan Wang, Le~Lu,
  Alan~L Yuille, and Yuyin Zhou,
\newblock ``Transunet: Transformers make strong encoders for medical image
  segmentation,''
\newblock {\em arXiv:2102.04306}, 2021.

\bibitem{hatamizadeh2022unetr}
Ali Hatamizadeh, Yucheng Tang, Vishwesh Nath, Dong Yang, Andriy Myronenko,
  Bennett Landman, Holger~R Roth, and Daguang Xu,
\newblock ``Unetr: Transformers for 3d medical image segmentation,''
\newblock in {\em WACV}, 2022.

\bibitem{xie2022simmim}
Zhenda Xie, Zheng Zhang, Yue Cao, Yutong Lin, Jianmin Bao, Zhuliang Yao,
  Qi~Dai, and Han Hu,
\newblock ``Simmim: A simple framework for masked image modeling,''
\newblock in {\em CVPR}, 2022.

\bibitem{chen2020improved}
Xinlei Chen, Haoqi Fan, Ross Girshick, and Kaiming He,
\newblock ``Improved baselines with momentum contrastive learning,''
\newblock {\em arXiv:2003.04297}, 2020.

\bibitem{chen2020simple}
Ting Chen, Simon Kornblith, Mohammad Norouzi, and Geoffrey Hinton,
\newblock ``A simple framework for contrastive learning of visual
  representations,''
\newblock in {\em ICML}, 2020.

\bibitem{he2020momentum}
Kaiming He, Haoqi Fan, Yuxin Wu, Saining Xie, and Ross Girshick,
\newblock ``Momentum contrast for unsupervised visual representation
  learning,''
\newblock in {\em CVPR}, 2020.

\bibitem{grill2020bootstrap}
Jean-Bastien Grill, Florian Strub, Florent Altch{\'e}, Corentin Tallec, Pierre
  Richemond, Elena Buchatskaya, Carl Doersch, Bernardo Avila~Pires, Zhaohan
  Guo, Mohammad Gheshlaghi~Azar, et~al.,
\newblock ``Bootstrap your own latent-a new approach to self-supervised
  learning,''
\newblock {\em NIPS}, 2020.

\bibitem{hua2021feature}
Tianyu Hua, Wenxiao Wang, Zihui Xue, Sucheng Ren, Yue Wang, and Hang Zhao,
\newblock ``On feature decorrelation in self-supervised learning,''
\newblock in {\em CVPR}, 2021.

\bibitem{ma2021abdomenct}
Jun Ma, Yao Zhang, Song Gu, Cheng Zhu, Cheng Ge, Yichi Zhang, Xingle An,
  Congcong Wang, Qiyuan Wang, Xin Liu, et~al.,
\newblock ``Abdomenct-1k: Is abdominal organ segmentation a solved problem?,''
\newblock {\em IEEE TPAMI}, 2021.

\bibitem{landman2015miccai}
Bennett Landman, Zhoubing Xu, J~Igelsias, Martin Styner, T~Langerak, and Arno
  Klein,
\newblock ``Miccai multi-atlas labeling beyond the cranial vault--workshop and
  challenge,''
\newblock in {\em MICCAI Workshop}, 2015.

\bibitem{oh2016deep}
Hyun Oh~Song, Yu~Xiang, Stefanie Jegelka, and Silvio Savarese,
\newblock ``Deep metric learning via lifted structured feature embedding,''
\newblock in {\em CVPR}, 2016.

\bibitem{he2016deep}
Kaiming He, Xiangyu Zhang, Shaoqing Ren, and Jian Sun,
\newblock ``Deep residual learning for image recognition,''
\newblock in {\em CVPR}, 2016.

\bibitem{ioffe2015batch}
Sergey Ioffe and Christian Szegedy,
\newblock ``Batch normalization: Accelerating deep network training by reducing
  internal covariate shift,''
\newblock in {\em ICML}, 2015.

\bibitem{huang2018decorrelated}
Lei Huang, Dawei Yang, Bo~Lang, and Jia Deng,
\newblock ``Decorrelated batch normalization,''
\newblock in {\em CVPR}, 2018.

\bibitem{huang2019iterative}
Lei Huang, Yi~Zhou, Fan Zhu, Li~Liu, and Ling Shao,
\newblock ``Iterative normalization: Beyond standardization towards efficient
  whitening,''
\newblock in {\em CVPR}, 2019.

\bibitem{mmselfsup2021}
MMSelfSup Contributors,
\newblock ``{MMSelfSup}: Openmmlab self-supervised learning toolbox and
  benchmark,'' {https://github.com/open-mmlab/mmselfsup}, 2021.

\bibitem{mmseg2020}
MMSegmentation Contributors,
\newblock ``{MMSegmentation}: Openmmlab semantic segmentation toolbox and
  benchmark,'' {https://github.com/open-mmlab/mmsegmentation}, 2020.

\bibitem{bardes2022vicregl}
Adrien Bardes, Jean Ponce, and Yann LeCun,
\newblock ``Vicregl: Self-supervised learning of local visual features,''
\newblock {\em NIPS}, 2022.

\bibitem{chen2018dcan}
Hao Chen, Xinjian Qi, Lequan Yu, and Pheng-Ann Heng,
\newblock ``Dcan: Deep contour-aware networks for accurate gland
  segmentation,''
\newblock {\em IEEE TMI}, 2018.

\end{thebibliography}

\end{document}